# Psychological Counseling Ability of Large Language Models

Fangyu Peng[1-5], Jingxin Nie[1-5]*

[1] Philosophy and Social Science Laboratory of Reading and Development in Children and Adolescents (South China Normal University), Ministry of Education Center for Studies of Psychological Application, South China Normal University; Guangzhou, 510631, China.

[2] Center for Studies of Psychological Application, South China Normal University; Guangzhou, 510631, China.

[3] Key Laboratory of Brain, Cognition and Education Sciences (South China Normal University), Ministry of Education.

[4] School of Psychology, South China Normal University; Guangzhou, 510631, China.

[5] Guangdong Key Laboratory of Mental Health and Cognitive Science, South China Normal University; Guangzhou, 510631, China.

*Corresponding author Email: niejingxin@gmail.com.

## Abstract

With the development of science and the continuous progress of artificial intelligence technology, Large Language Models (LLMs) have begun to be widely utilized across various fields. However, in the field of psychological counseling, the ability of LLMs have not been systematically assessed. In this study, we assessed the psychological counseling ability of mainstream LLMs using 1096 psychological counseling skill questions which were selected from the Chinese National Counselor Level 3 Examination, including Knowledge-based, Analytical-based, and Application-based question types. The analysis showed that the correctness rates of the LLMs for Chinese questions, in descending order, were GLM-3 (46.5%), GPT-4 (46.1%), Gemini (45.0%), ERNIE-3.5 (45.7%) and GPT-3.5 (32.9%). The correctness rates of the LLMs for English questions, in descending order, were ERNIE-3.5 (43.9%), GPT-4 (40.6%), Gemini (36.6%), GLM-3 (29.9%) and GPT-3.5 (29.5%). A chi-square test indicated significant differences in the LLMs' performance on Chinese and English questions. Furthermore, we subsequently utilized the Counselor's Guidebook (Level 3) as a reference for ERNIE-3.5, resulting in a new correctness rate of 59.6%, a 13.8% improvement over its initial rate of 45.8%. In conclusion, the study assessed the psychological counseling ability of LLMs for the first time, which may provide insights for future enhancement and improvement of psychological counseling ability of LLMs.

## Keywords

psychological counseling, Large Language Models, assessment

## 1. Introduction

In recent years, Large Language Models (LLMs) have made significant progress in the field of natural language processing (Guu et al., 2020). Through pre-training and fine-tuning techniques (Chowdhery et al., 2023), LLMs are able to comprehend and follow human commands, thus demonstrating superior performance in a wide range of tasks (Wei et al., 2022). LLMs such as ChatGPT are rapidly changing human interaction with AI and raising questions about the nature of human intelligence and consciousness (Hintze, 2023). While LLMs were not designed to explicitly capture or mimic elements of human cognition and psychology, recent research suggests that LLMs may have spontaneously developed this ability due to the large pool of human-generated language content they have been trained with. For example, LLMs display properties similar to human cognitive and emotional abilities and processes, including theory of mind (Kosinski, 2023), emotional intelligence (Wang et al., 2023), emotion recognition (Elyoseph et al., 2024), heuristic decision making (Suri et al., 2024), cognitive biases in decision making (Hagendorff, Fabi & Kosinski, 2023), mental trait inference (Peters & Matz, 2024), quantitative and verbal ability (Hickman, Dunlop & Wolf, 2024), and semantic priming (Digutsch & Kosinski, 2023).

Psychological counseling refers to the use of psychological principles and methods by professionally trained counselors to help-seekers discover their own problems and their root causes, so as to tap the potential ability of the seekers themselves to change the original cognitive structure and behavioral patterns, in order to improve their adaptability to life and their ability to regulate the surrounding environment (Ma, 2014). In modern society, as more and more people are in a state of sub-optimal mental health(Huang & Tang, 2021), the demand for psychological counselors is also increasing. However, the psychological counseling market lacks a comprehensive training mechanism and system, making it difficult to cultivate psychological counseling talents. Furthermore, the high costs may deter potential candidates from pursuing this field (Liu, 2023). Secondly, psychological counselors often have multiple careers, which leads to difficulties in shifting their identities when dealing with visitors, making it hard to maintain objectivity in counseling. Moreover, traditional counseling usually involves appointments, about 1 to 2 times a week, which may not ensure the timeliness of psychological counseling for visitors (Liu, 2023). On the other hand, the stigmatization of mental illnesses has caused some people with mental illnesses to be ashamed to admit that they suffer from mental illnesse.

The emergence of LLMs makes it possible for them to act as virtual counselors, avoiding the disadvantages of traditional counseling. LLMs can maintain the objectivity and neutrality of the subject in counseling, and their covert and anonymous nature can also reduce the psychological defenses of visitors. LLMs are more real-time and can provide timely psychological guidance and feedback to visitors anytime, anywhere. Research has shown that LLMs can help patients understand and cope with psychological difficulties, such as interventions for patients with depression (Chen & Yan, 2024). LLMs can also be used to understand suicide by conducting a linguistic analysis of Internet users' discussions on suicide-related topics (Bauer et al., 2024), and they can obtain information about users' expressions, emotional states, and concerns from online texts and comments, generating a comprehensive and objective mental health assessment report (Chen & Yan, 2024). Thus, psychological counseling is a potential and promising application area for LLMs.

Though a few LLMs models for counseling, such as EmoGPT (Liu, 2023), SoulChat (Chen et al., 2023) and MentaLLaMA (Yang et al., 2024), have been conducted, there are no studies that have systematically and quantitatively assessed the counseling competencies of LLMs. LLMs are difficult to quantitatively assess in terms of efficacy, completion of counseling goals, and other metrics when acting as psychological counselors. Visitors may not consistently use LLMs for counseling, making it difficult to collect enough data to assess long-term efficacy. Furthermore, most of the training data used in the mental health large models are conversational data, psychological theories, and so on. There is a discrepancy between these training data and actual practice. In practice, it is a challenge for the model to be flexible in applying these theories to the specific situation of the visitor.

As shown in Figure 1, this study aims to assess the psychological counseling competence of LLMs and provide insights for future enhancement of LLMs' psychological counseling competence. First, an objective assessment dataset was established, which was original designed for human psychological counselors qualification test. Then, this dataset was used to assess the existing mainstream LLMs, including ERNIE-3.5, GLM-3, Gemini, GPT-3.5 and GPT-4. Since mainstream LLMs are not specifically trained for psychological counseling, they may lack the relevant professional knowledge in this field. Therefore, we considered using Retrieval-Augmented Generation (RAG) technology to provide the model with specialized domain knowledge to see if it could enhance its psychological counseling capabilities.

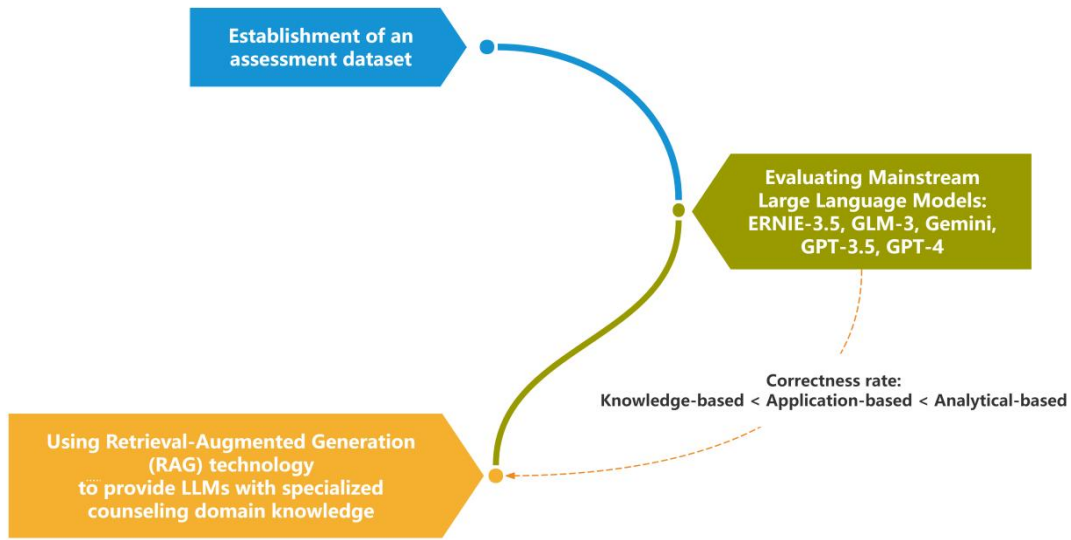

Figgure.1 Flow chart of assessing psychological counseling competence in the LLMs

## 2. Methods

### 2.1. Assessment Dataset

The assessment dataset includes 1096 psychological counseling skills questions that were selected from the Chinese National Counselor Level 3 Exam. The Ministry of Human Resources and Social Security of China officially certifies the exam, which is divided into a theoretical knowledge test and a skill operation assessment part. Both parts are written exams, graded on a hundred-mark system, where a score of 60 or above is considered passing. Passing the examination was a necessary condition for becoming a national professional qualification psychological counselor. The skill operation assessment part contains skill choice questions and case quiz questions, of which 80% of the points are allocated to skill choice questions and 20% to case quiz questions (Xu, 2011). The test questions selected for the assessment dataset are the skill choice questions of the skill operation assessment part, as this part not only contains theoretical knowledge, but is also closer to practical application, and it is easier to objectively judge the correctness of the choice questions. The skill choice question comprises cases and choice questions, each with four alternative options. For single choice questions, only one option is correct; for multiple choice questions, two or more options are correct, and selecting wrong, fewer, or more than one options is considered as incorrect.

These questions also can be categorized as three types: Knowledge-based, Analytical-based and Application-based (some examples are shown in Table 1), consisting of 192, 455 and 449 questions, respectively. The Knowledge-based questions mainly examine the degree of mastery of basic knowledge, including definitions, characteristics, classifications, theories, and so on. They typically necessitate direct responses grounded in existing knowledge, devoid of intricate analysis or application. The Analysis-based questions require an in-depth analysis and understanding of an issue or phenomenon based on a case or a conversation, involving an exploration of symptoms, subconsciousness, causes, relationships, and so forth. The third type of questions, Application-based questions, integrates characteristics of both Knowledge-based and Analysis-based questions. It centers on examining the ability to apply existing knowledge to real cases. This not only requires a thorough understanding and mastery of pertinent knowledge by LLMs but also a keen insight and accurate analysis ability of psychological counseling cases.

Table 1. Question Types for Assessment

| Question Types | n (%) | Examples |
|---|---|---|
| Knowledge-based | 192（17.5） | Breathing relaxation method does not include ( ). |
| | | Common methods of behavior modification include ( ). |
| Analytical-based | 455（41.5） | The help-seeker's current emotional symptoms include ( ). |
| | | Possible causes of the help-seeker's psychological problems include ( ). |
| | | The help-seeker's personality traits do not include ( ). |
| Application-based | 449（41.0） | The most likely diagnosis for this help-seeker is ( ). |
| | | Influential techniques used by the psychological counselor during the counseling process include ( ). |

## 2.2. Assessment Procedure

A variety of mainstream LLMs, including GPT-4, GPT-3.5, GLM-3, Gemini, and ERNIE-3.5, were evaluated by the Counselling skills test. Given that these models were trained on English and/or Chinese datasets, two versions of the questions were used for testing, i.e., Chinese and English, to more accurately assess their performance and capabilities. The two versions of the prompts are presented in Table 2.

Table 2. Assessment Prompts for LLMs

| Versions | Prompts |
|---|---|
| Chinese version | 单选题要求只选择一个正确的选项，多选题要求选择两个或两个以上的正确的选项。假如你是一名心理咨询师，请根据三个引号内案例或对话内容、题号后单选或多选的要求，回答案例或对话后的单选或多选题，不用解释或解析，回答形式为题号、选项，如 1、B。 |
| English version | Only one correct choice is selected for single choice, and two or more correct choices are selected for multiple choice. If you are a psychological counselor, please answer the single or multiple choice questions after the case or conversation based on the content of the case or conversation, without explanation or analysis, in the form of question numbers and options, like 1: B. |

## 2.3. Retrieval-Augmented Generation for Psychological Counseling

Since mainstream large language models are not specifically trained for psychological counseling, they may lack the relevant professional knowledge in this field. Therefore, we consider using Retrieval-Augmented Generation (RAG) technology to provide the model with specialized domain knowledge to see if it can enhance its psychological counseling capabilities. Retrieval-Augmented Generation (RAG) was proposed by Meta AI in 2020, and its basic idea is to combine retrieval systems with generative models to integrate knowledge in a modular way (Lewis et al., 2020). In our experiment, the Psychological Counselor's Guidebook (Ji, 2016) was provided as reference to ERNIE-3.5 before it answers the questions.

**2.4. Statistical analysis**

The data analysis was conducted using SPSS 24.0. First, the correctness rate of different types of questions and different LLMs were calculated. Next, the chi-square test was conducted to test if there is any different of correctness between types of questions and LLMs. Subsequently, after using the RAG technique, the new rate of correctness rates were calculated and compared to original correctness by chi-square test.

# 3. Results

## 3.1. Accuracy Characteristics of LLMs in Question Answering Tasks

As shown in Table 3, for the three Chinese types of questions, the total correctness rates of all five LLMs (ERNIE-3.5, GLM-3, Gemini, GPT-3.5 and GPT-4) were 40.8%, 44.8% and 42.9%, respectively. In contrast, for the same types of English questions, the total correctness rates were 33.6%, 38.5% and 34.7%, respectively.

Three one-way chi-square tests were conducted to assess the differences in the accuracy of the answers among the three types of test questions in the Chinese and English versions (see Table 3). The results showed that the difference in the correctness rates between the two language versions was statistically significant in all three types of questions ($\chi^2_1$ =10.61, $df_1$ = 1, $p_1$ = 0.001; $\chi^2_2$ = 18.23, $df_2$ = 1, $p_2$ < 0.001; ($\chi^2_3$ = 31.42, $df_3$ = 1, $p_3$ < 0.001). The chi-square test was further followed by multiple comparisons to assess the correctness of the answers on the two languages for each of the three question types. The results showed that there was a significant difference between the correctness rates of the questions for the knowledge-based, analytical-based and application-based questions respectively on the two language versions. Regardless of the question types, the Chinese version had significantly higher correctness rates than the English version.

A one-way two-level chi-square test was conducted to assess the correctness of LLMs in answering questions presented in two language versions (Chinese and English) (see Table 3). The results indicated that the difference in the percentage of correctness between the two language versions was statistically significant ($\chi^2$ = 59.16, $df$ = 1, $p$ < 0.001). The chi-square test was further followed by multiple comparisons to assess the correctness of the LLMs' answers in both language versions of the questions. The results showed a significant difference in correctness rates between the English and Chinese versions of the questions, with the Chinese version showing significantly higher correctness rates than the English version.

Table 3. Analysis of the Characteristics of Correctness of three types questions in different languages and Correctness of two language versions

| Question Types | Test version | | $\chi^2$ | $p$ |
|---|---|---|---|---|
| | Chinese version | English version | | |
| Knowledge-based | 0.408[a] | 0.336[b] | 10.61 | 0.001 |
| Analytical-based | 0.448[a] | 0.385[b] | 18.23 | < 0.001 |
| Application-based | 0.429[a] | 0.347[b] | 31.42 | < 0.001 |
| All types | 0.433[a] | 0.361[b] | 59.16 | < 0.001 |

Note: $\chi2$ test = chi-square test; Significance tests for multiple comparisons were performed using the letter labeling method, where the same labeled letter indicated that there was no difference between the corresponding two sets of data, and a different letter indicated that the difference was statistically significant.

Correctness of different LLMs was reported in Table 4. The correctness rates of the LLMs for Chinese version were as follows: 46.5% for GLM-3, 46.1% for GPT-4, 45.8% for ERNIE-3.5, 45.0% for Gemini and 32.9% for GPT-3.5. The correctness rates of the LLMs for the English version were as follows: ERNIE-3.5, GPT-4, Gemini, GLM-3, GPT-3.5, with the correctness rate of 43.9%, 40.6%, 36.6%, 29.9% and 29.5%, respectively.

A one-way five-level chi-square test chi-square test was conducted on the five LLMs (ERNIE-3.5, GLM-3, Gemini, GPT-3.5 and GPT-4) for doing Chinese questions correctly, and the results showed that the difference in the correctness rates of the five LLMs for doing Chinese questions was statistically significant ($\chi^2 = 59.91$, $df = 4$, $p < 0.001$). The chi-square test was further followed by multiple comparisons of the correctness of different LLMs in doing Chinese questions, and the results showed that ERNIE-3.5, GLM-3, Gemini and GPT-4 were significantly higher than the correctness of GPT-3.5 in doing Chinese questions.

A one-way five-level chi-square test was also conducted on the five LLMs (ERNIE-3.5, GLM-3, Gemini, GPT-3.5 and GPT-4) for correctly doing English questions, and it was found that the difference between the five LLMs in terms of correctly doing English questions was statistically significant ($\chi^2 = 77.54$, df = 4, p < 0.001). The chi-square test was further followed by multiple comparisons of the correctness of different LLMs in doing English questions, and ERNIE-3.5's correctness in doing English questions was significantly higher than that of GLM-3, Gemini and GPT-3.5. The correctness rate of Gemini doing English questions is significantly higher than the correctness rate of GLM-3 and GPT-3.5 in English version.

Table 4. Analysis of the Characteristics of Correctness in Question Answering Tasks Performed by Different LLMs

| Test version | LLM | Correct | Correctness rate | $\chi^2$ | $p$ |
|---|---|---|---|---|---|
| Chinese version | ERNIE-3.5 | 502[b] | 0.458 | | |
| | GLM-3 | 510[b] | 0.465 | | |
| | Gemini | 493[b] | 0.450 | 60.11 | < 0.001 |
| | GPT-3.5 | 361[a] | 0.329 | | |
| | GPT-4 | 505[b] | 0.461 | | |
| | GLM-3, GPT-4, ERNIE-3.5, Gemini > GPT-3.5 ($p < 0.05$) | | | | |
| English version | ERNIE-3.5 | 481[a] | 0.439 | | |
| | GLM-3 | 328[c] | 0.299 | | |
| | Gemini | 401[b] | 0.366 | 77.54 | < 0.001 |
| | GPT-3.5 | 323[c] | 0.295 | | |
| | GPT-4 | 445[a,b] | 0.406 | | |
| | ERNIE-3.5 > Gemini, GLM-3, GPT-3.5; Gemini > GLM-3, GPT-3.5 ($p < 0.05$) | | | | |

Note: LLM = Large Language Model; χ2 test = chi-square test; Significance tests for multiple comparisons were performed using the letter labeling method, where the same labeled letter indicated that there was no difference between the corresponding two sets of data, and a different letter indicated that the difference was statistically significant.

## 3.2. Accuracy Characteristics of LLMs in Different Types of Questions

Five one-way three-level chi-square tests were conducted to evaluate the correctness of ERNIE-3.5, GLM-3, Gemini, GPT-3.5, and GPT-4 when answering three types of Chinese questions (Knowledge-based, Analytical-based and Application-based). The results in Table 5 indicate that the differences in accuracy rates among these five LLMs when answering three types of Chinese questions were not statistically significant.

Separate one-way three-level chi-square tests were performed for ERNIE-3.5, GLM-3, Gemini, and GPT-4 to assess their correctness in answering three types of English questions (Knowledge-based, Analytical-based and Application-based). The results showed that the differences in correctness among three LLMs when answering English questions were also not statistically significant. However, when a one-way three-level chi-square test was conducted on GPT-3.5's correctness in answering the three types of English questions, it was found that the difference in GPT-3.5's correctness was statistically significant ($\chi^2$ = 6.60, $df$ = 2, $p$ = 0.037 < 0.05). The chi-square test was further followed by multiple comparisons of GPT-3.5's correctness in doing the three types of English questions, and the results showed that GPT-3.5's correctness in doing Knowledge-based was significantly lower than the correctness in doing Analytical-based English questions.

Table 5. Analysis of the Characteristics of Correctness of LLMs in Answering Different Types

| Test version | LLM | Type of question | Correct | Correctness rate | $\chi^2$ | $p$ |
|---|---|---|---|---|---|---|
| Chinese version | ERNIE-3.5 | Knowledge-based | 84 | 0.438 | | |
| | | Analytical-based | 211 | 0.464 | 0.40 | 0.82 |
| | | Application-based | 207 | 0.461 | | |
| | GLM-3 | Knowledge-based | 87 | 0.453 | | |
| | | Analytical-based | 225 | 0.495 | 2.55 | 0.28 |
| | | Application-based | 199 | 0.443 | | |
| | Gemini | Knowledge-based | 82 | 0.427 | | |
| | | Analytical-based | 214 | 0.470 | 1.31 | 0.52 |
| | | Application-based | 198 | 0.441 | | |
| | GPT-3.5 | Knowledge-based | 53 | 0.276 | | |
| | | Analytical-based | 150 | 0.330 | 3.50 | 0.17 |
| | | Application-based | 158 | 0.352 | | |
| | GPT-4 | Knowledge-based | 86 | 0.448 | | |
| | | Analytical-based | 219 | 0.481 | 1.33 | 0.52 |
| | | Application-based | 200 | 0.445 | | |
| | ERNIE-3.5 (RAG) | Knowledge-based | 124 | 0.646 | | |
| | | Analytical-based | 311 | 0.684 | 0.96 | 0.62 |
| | | Application-based | 298 | 0.664 | | |
| English version | ERNIE-3.5 | Knowledge-based | 86 | 0.448 | | |
| | | Analytical-based | 209 | 0.459 | 1.75 | 0.42 |
| | | Application-based | 187 | 0.416 | | |
| | GLM-3 | Knowledge-based | 55 | 0.286 | | |
| | | Analytical-based | 147 | 0.323 | 2.12 | 0.35 |
| | | Application-based | 126 | 0.281 | | |
| | Gemini | Knowledge-based | 64 | 0.333 | | |
| | | Analytical-based | 183 | 0.402 | 4.48 | 0.11 |
| | | Application-based | 154 | 0.343 | | |
| | GPT-3.5 | Knowledge-based | 42[b] | 0.219 | | |
| | | Analytical-based | 144[a] | 0.316 | 6.60 | 0.04 |
| | | Application-based | 137[a] | 0.305 | | |
| | GPT-4 | Knowledge-based | 76 | 0.396 | | |
| | | Analytical-based | 194 | 0.426 | 1.36 | 0.51 |
| | | Application-based | 175 | 0.390 | | |

Note: LLM = Large Language Model; χ2 test = chi-square test; Significance tests for multiple comparisons were performed using the letter labeling method, where the same labeled letter indicated that there was no difference between the corresponding two sets of data, and a different letter indicated that the difference was statistically significant.

### 3.3. Assessment of LLMs with Psychological Counselor's Guidebook

To improve the accuracy of LLM's responses to questions, ERNIE-3.5 used the Psychological Counselor's Guidebook published by China Machine Press as a reference document (Ji, 2016). It re-examined the questions in Chinese and analyzed them statistically accordingly. The results of chi-square test showed that the correctness of ERNIE-3.5 questions was significantly different before and after the use of the counselor's guidebook ($\chi^2 = 41.73$, $df = 1$, $p < 0.001$). The correctness of the questions after referring the document was significantly higher than the correctness of the questions previously done. The percentage of questions done correctly without referencing the document was 45.8%, which increased to 59.6% after using the document, representing an improvement of 13.8%.

Three one-way two-level chi-square tests were conducted to assess ERNIE-3.5's correctness before and after using guidebook, categorized by question type. As shown in Table 6, the differences in correctness between ERNIE-3.5's responses before and after using guidebook for each of the three question types (Knowledge-based, Analytical-based and Application-based) were statistically significant ($\chi^2_1 = 9.38$, $df_1 = 1$, $p_1 = 0.002 < 0.05$; $\chi^2_2 = 19.25$, $df_2 = 1$, $p_2 < 0.001$; $\chi^2_3 = 13.50$, $df_3 = 1$, $p_3 < 0.001$). The chi-square test was further followed by multiple comparisons of correctness of ERNIE-3.5 before and after prompting, showed that the correctness rate after using guidebook was significantly higher than that before prompting, regardless of the type of question.

Table 6. Analysis of ERNIE-3.5's Correctness Characteristics Before and After Incorporating Counselors' Guide book

| Type of question | RAG | Correct | Correctness rate | $\chi^2$ | $p$ |
|---|---|---|---|---|---|
| Knowledge-based | pre-enhancement | 84[a] | 0.438 | 9.38 | 0.002 |
| | post-enhancement | 114[b] | 0.594 | | |
| Analytical-based | pre-enhancement | 211[a] | 0.464 | 19.25 | < 0.001 |
| | post-enhancement | 277[b] | 0.609 | | |
| Application-based | pre-enhancement | 207[a] | 0.461 | 13.50 | < 0.001 |
| | post-enhancement | 262[b] | 0.584 | | |
| Chinese Questions | pre-enhancement | 502[a] | 0.458 | 41.73 | < 0.001 |
| | post-enhancement | 653[b] | 0.596 | | |

Note: LLM = Large Language Model; χ2 test = chi-square test; Significance tests for multiple comparisons were performed using the letter labeling method, where the same labeled letter indicated that there was no difference between the corresponding two sets of data, and a different letter indicated that the difference was statistically significant.

## 4. Discussion

In this study, we used 1096 counseling skills questions selected from the test questions of the Chinese National Counselor Level 3 Exam to assess the psychological counseling competence of mainstream LLMs through their question answering. Our findings indicate that the tested LLMs exhibited relatively low accuracy in the assessment, with an average of 43.26% for Chinese questions and 36.10% for English questions, suggesting their psychological counseling capabilities could be improved significantly.

Large language model (LLM) demonstrated varying performance across different languages, with the correct rate for Chinese questions generally being higher than that for English questions. This may be attributed to the fact that some LLMs were more heavily optimized for the Chinese model during their development and training phases, resulting in a higher accuracy for Chinese questions compared to English ones. Meanwhile, it is possible that the questions, originally in Chinese, undergo some bias when translated into English, which could affect the LLM's understanding and subsequent performance on the English questions. The analysis concluded that, notable disparities in correctness rates were observed among different LLMs when tackling both Chinese and English questions, underscoring the varying linguistic comprehension and processing abilities among these models.

There are three types of questions in this study, Knowledge-based, Analytical-based and Application-based. The corresponding types of errors can also be categorized into Knowledge-based, Analytical-based and Application-based errors. Knowledge-based errors refer to factual errors or conceptual confusion in the answers due to inaccurate understanding or incomplete mastery of relevant knowledge during the answering process. Analytical-based errors refer to deviations or errors in the analysis results due to improper logical reasoning, unclear analytical thinking or failure to accurately grasp the core of the problem in the process of analyzing the problem in combination with the case or dialogue. Application-based errors refer to the failure to get the correct answer to a question when applying relevant counseling knowledge or skills to solve a multiple-choice question following a case or a dialogue due to an incomplete grasp of the knowledge points or a failure to flexibly apply the relevant knowledge. The study concluded that LLMs performed relatively better on Analytical-based questions (44.8% in Chinese and 38.5% in English), but did not show an advantage on Knowledge-based questions (40.8% in Chinese and 33.6% in English). This may imply that LLMs are more adept at understanding and analyzing information in known situations and have limitations in theoretical knowledge of counseling psychology. In addition, the performance of Application-based questions (42.9% in Chinese and 34.7% in English) was in between, indicating that LLMs are competent in applying their knowledge to case situations, but there is still room for improvement. In general, the differences between the types of questions were not significant, and LLMs did not get more than 50% correct on all types of questions, indicating that although LLMs have made significant progress in the field of natural language processing, there are still major limitations in comprehending and answering complex questions.

Since mainstream LLMs are not specifically trained for psychological counseling, they tend to yield lower accuracy rates when answering Knowledge-based questions pertinent to this domain. Therefore, we used Retrieval-Augmented Generation (RAG) technology to see if it could enhance its psychological counseling capabilities. After ERNIE-3.5 was provided with the Psychological Counselor's Guidebook for reference, its correctness rate improved, rising from 45.8% to 59.6%. The correctness rate of questions done after the reference is significantly higher than that before the reference, demonstrating LLMs robust learning and adaptation capabilities within the realm of psychological counseling. This advancement predicted its potential to pass the Level 3 psychological counselor examination and underscores the capacity of LLMs to excel in standardized tests, potentially even qualifying them as professional psychological counselors. The improvement also pinpoints the lack of domain-specific knowledge as a crucial factor limiting their performance. With the continued advancement of technology and the accumulation of data, LLMs can be increasingly fine-tuned and optimized for specific domains such as mental health and psychological counseling. In summary, our research marks the first comprehensive measurement and evaluation of the psychological counseling ability of LLMs, potentially offering insights for future enhancements and refinements of their capability in this field.

This study provides a new assessment of LLM counseling competence, but several limitations remain. First, we only examined some of the mainstream LLMs, namely GPT-4, GPT-3.5, GLM-3, Gemini, and ERNIE-3.5. Despite the popularity of these LLMs, many other LLMs exist, particularly the Mental Health Large Model. Future LLMs may perform better on counseling aptitude tests. Second, this study assessed LLMs' counseling competence through three types of questions: Knowledge-based, Analytical-based, and Application-based, which only provided a limited perspective, but the actual counseling process is more than that. Knowledge of counseling theory does not necessarily reflect the model's ability to practice in actual counseling, and since the theory and practice of counseling is constantly evolving, the model may not be able to update itself in real time with the latest research findings and therapeutic approaches. Although a number of real-life examples appeared in the case study questions, the model can only analyze based on predetermined information, and is not able to engage in real-life conversations and adjustments. Future research could benefit from larger and more diverse tests, and more scenarios could be designed to examine whether LLMs are able to counsel effectively. Finally, the prompts in this study used only simple direct questions and answers, which may affect the reliability and accuracy of the results generated by LLMs.

## Funding

This work was supported by the Research Center for Brain Cognition and Human Development, Guangdong, China (No. 2024B0303390003); and the Striving for the First-Class, Improving Weak Links and Highlighting Features (SIH) Key Discipline for Psychology in South China Normal University.